\documentclass{article}

    \PassOptionsToPackage{numbers, compress}{natbib}

\usepackage{neurips_2021}

\usepackage[utf8]{inputenc} 
\usepackage[T1]{fontenc}    
\usepackage{hyperref}       
\usepackage{url}            
\usepackage{booktabs}       
\usepackage{amsfonts}       
\usepackage{nicefrac}       
\usepackage{microtype}      
\usepackage{xcolor}         
\usepackage{graphicx}
\usepackage{subfigure}
\usepackage{amsmath, bm}
\title{A Method For Eliminating Contour Errors In Self-Encoder Reconstructed Images}

\author{@
  David S.~Hippocampus\thanks{Use footnote for providing further information
    about author (webpage, alternative address)---\emph{not} for acknowledging
    funding agencies.} \\
  Department of Computer Science\\
  Cranberry-Lemon University\\
  Pittsburgh, PA 15213 \\
  \texttt{hippo@cs.cranberry-lemon.edu} \\
}

\begin{document}

\maketitle

\begin{abstract}
  In industry, products in the same product line are mostly similar. It is often a normal idea to use a self-encoder to reconstruct the image\cite{ref1, ref2} in order to find product defects or foreign bodies in a differential way\cite{ref3, ref4}. In general, however, the edge errors in the reconstructed images are often large\ref{fig:pad_p2}. No previous papers have proposed solutions to this problem. The authors observe a priori that two similar reconstructed networks (similar structure, different parameters, same training method) with the same image, with the output results subtracted, can obtain stable image edge information without foreign objects or noise, even if the input image has foreign objects. In this paper, we propose a self-supervised twin network approach based on this a priori. The method of generating the approximate edge information of an image and then differentially eliminating the edge errors in the reconstructed image with a dilate algorithm. This is used to improve the accuracy of the reconstructed image and to separate foreign matter and noise from the original image, so that it can be visualized in a more practical scene.
\end{abstract}

\section{Introduction}

Surface defects have a huge impact on industrial products and their quality, and in industrial inspection, we often deploy models to edge devices, where computing resources are scarce but real-time requirements are high. At the same time, external factors such as environment, background and illumination increase the difficulty of detection. This has led to higher requirements for defect detection in a wide range of practical industrial applications. Over the past few years, researchers have been investigating computer vision-based defect detection and have achieved some results. Traditional image processing algorithms are based on statistical machine learning methods, which typically include histogram equalisation, filtering and denoising, grey-scale binarisation, re-filtering, etc. to obtain simple image information with front-to-back separation, followed by the labelling and detection of defects using mathematical morphology, Fourier variation and other algorithms and machine learning models.
\begin{figure}
\includegraphics[scale=0.25]{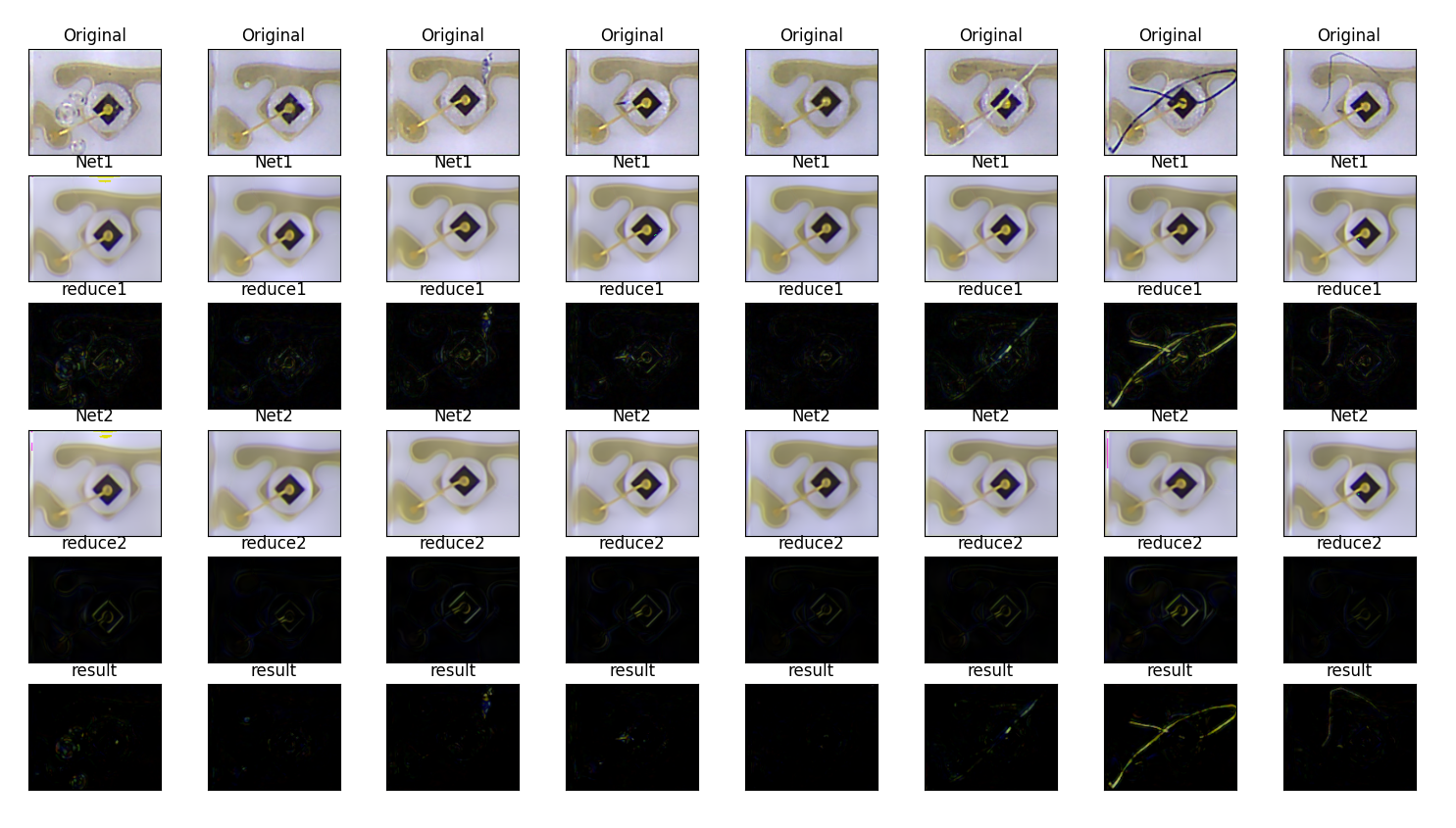}
\caption{A visualization of our proposed algorithm applied to foreign object detection is shown. The first row of images is the original to be detected and the last row of images is the output result of the algorithm. The middle row is a display of the output results of some intermediate steps.} 
\label{fig:1}
\end{figure}
The above algorithms have achieved good results in some specific applications but still have many shortcomings; the image processing steps are numerous and strongly targeted, the robustness is poor, and many algorithms are computationally intensive and cannot accurately detect the size and shape of defects, whereas deep learning can update the model weights directly by learning the image data, avoiding the complex algorithmic process of manual design. It is also extremely robust and accurate.
\begin{figure}[t]
\includegraphics[scale=0.08]{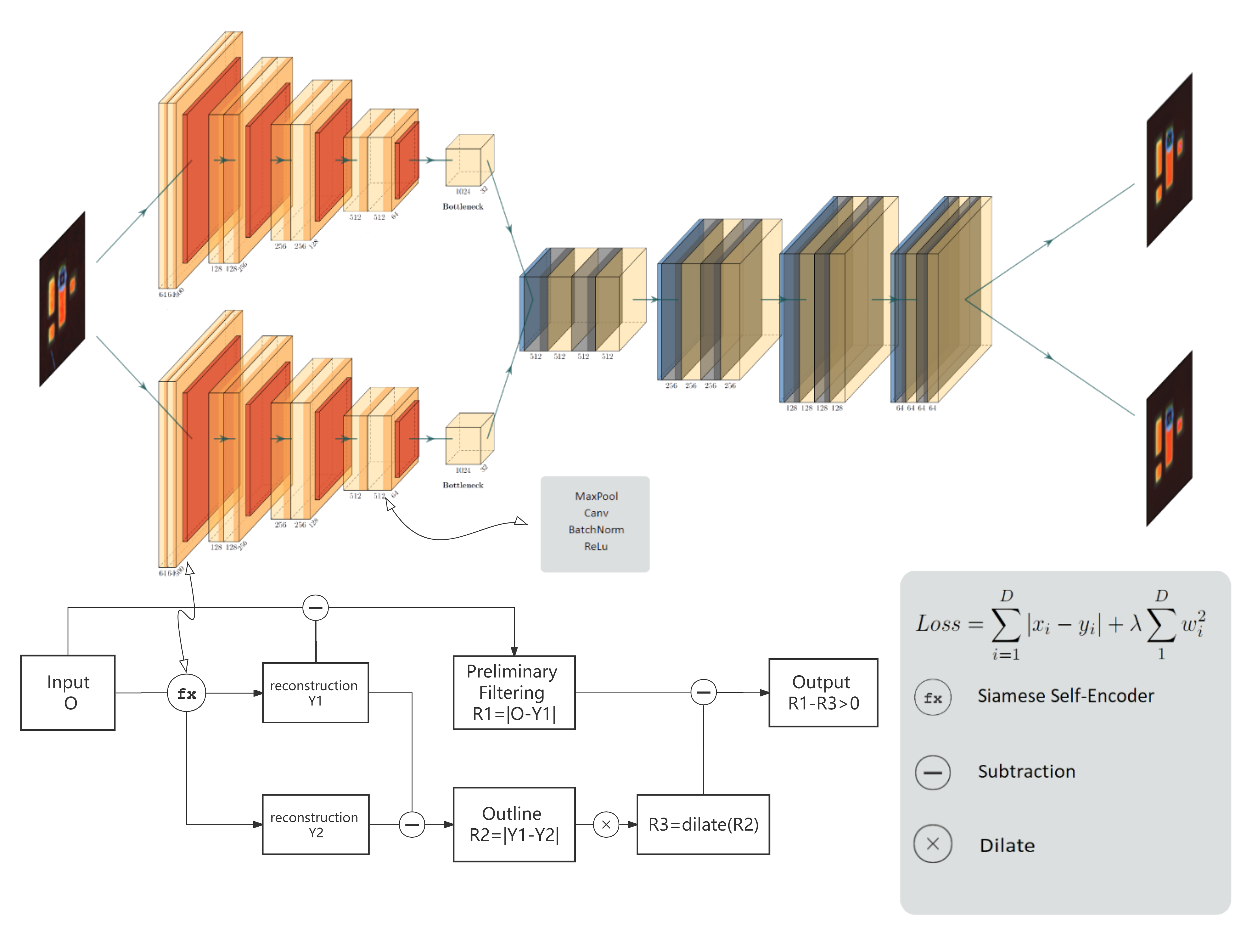}
\caption{The overall flow chart of the method. where \textbf{fx} can be replaced by a self-encoder network of any structure. In this paper we use this structure.} 
\label{fig:2}
\end{figure}
Due to the following problems with industrial data, 1) the data is commercially valuable and confidential, and the published data set is relatively small. 2) the sample distribution is uneven, i.e. the number of positive samples is much larger than the number of negative samples. This makes it difficult to use the relevant data to learn defective features well. 3) The collection and labelling of defects and foreign objects not only requires a lot of time and manpower but also a high level of a priori knowledge for the personnel involved. Based on the above problems semi-supervised and unsupervised learning is now a hot topic in defect detection. This shows that unsupervised networks are particularly suitable for defect and foreign object detection in industrial scenarios. It is of great value in the industrial field.

In this paper, we propose an unsupervised Siamese network that generates the approximate edge information of an image and then differentially eliminates the edge errors in the reconstructed image with an inflation algorithm as a way to improve the accuracy of the reconstructed image, separating foreign matter and noise from the original image, and by self-encoding the building blocks of two similar reconstructed networks (similar structure, different parameters, same training method), the output results are subtracted and can obtain Stable image edge information without foreign objects or noise, even if the input image has foreign objects.

\begin{figure}[htb]
\centering
\subfigure
[Input] 
{\includegraphics[scale=1]{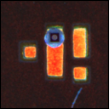}} 
\subfigure[Reconstruction]{\includegraphics[scale=1]{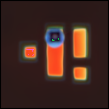}}
\subfigure[The Edge Error]{\includegraphics[scale=1]{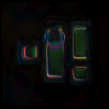}\label{fig:pad_p2}}

\caption{A single self-encoder inputs an image(a). Reconstruction produces a single image(b). Finally the two images are subtracted to get the edge error(c).} 
\label{fig:3}  
\end{figure}

The key contributions of this paper are :
\begin{itemize}
    \item Find a general method for unsupervised elimination of edge errors\ref{fig:pad_p2} in reconstructed images.
    \item We experimented that the method can be used for unsupervised foreign body detection. Only a sufficient number of positive samples are required, and even a small number of negative samples are allowed to be mixed with the positive samples.
    \item We verified the scope of application of the algorithm.
    \item We experiment with the effect of various training methods on the performance of such self-supervision.
\end{itemize}

The rest of the paper follows. In Section 3, we discuss the motivation for this idea. In Section IV we briefly review the recent advances in single classification models and self-encoder noise reduction. Then in Section IV, the procedure of the new approach proposed in this paper is described in detail, and in Section V, the performance bounds of the proposed approach are analyzed, as well as some experiments, and finally our subsequent work is given.

\section{Motivation}
\label{gen_inst}
Unsupervised self-coding networks were particularly suitable for the detection of product defects and foreign objects in industrial scenarios. Because these defects and foreign objects were often unpredictable and morphologically distributed, it was difficult to use the relevant data for supervised learning. Furthermore, collecting and labelling defects and foreign bodies was not only costly in terms of time and manpower but also requires a high level of a priori knowledge from the personnel involved. With a relatively easy to obtain the positive sample and a small number of occurrences of negative samples. Active learning of cured information from positive sample distributions 
could be helpful in determining the presence of foreign objects.

We had this very simple assumption that a picture to be detected is made up of a clean picture + noise + foreign matter together. Self-encoders were usually easy and did not require labelling to reconstruct the original image well. However, we suspected that this may be due to the presence of regular terms, or that the neural network is only a first-order Taylor approximation of the real image. Therefore, the reconstruction of the image is often accompanied by a sharp edge gradient error. This was reflected in the image as shown in Figure 1.

There is such a strong error. Then for most small foreign bodies, defects and anomalies, it can be difficult to separate positive and negative samples using differencing. Therefore, eliminating the edge gradient error becomes a hurdle that has to be crossed. We found a priori that could roughly find the location of edges that did not contain noise, foreign matter. Because it is an approximate location, we need to use dilate on it to wrap around the error location. After 3 times the difference algorithm, the noise, foreign matter information of the image can be extracted relatively well.

\section{Related work}
\label{headings}
In recent years, neural networks have achieved great results in computer vision applications, and these fields include target detection, face recognition, autonomous driving, etc. However, in the traditional industrial field, the application of deep neural networks is still difficult to land, mainly because the industrial field has high requirements for real-time and accuracy, so pure unsupervised foreign object recognition is still the mainstream solution.

Such as Single Class Support Vector Machine (SVDD) \cite{ref10}, OneClass SVM \cite{ref12}, OneClass MPM \cite{ref13}, using different statistical one class classifiers to get decision bounds, DeepSVDD \cite{ref8} to let then use deep neural networks to achieve a minimum size hypersphere to enclose the training data with an objective function of the following equation. 

\begin{displaymath}\min _{\mathcal{W}} \frac{1}{n} \sum_{i=1}^{n}\left\|\phi\left(\boldsymbol{x}_{i} ; \mathcal{W}\right)-\boldsymbol{c}\right\|^{2}+\frac{\lambda}{2} \sum_{\ell=1}^{L}\left\|\boldsymbol{W}^{\ell}\right\|_{F}^{2}\end{displaymath}

There are also works that use conditional generative adversarial networks to foreign labeled images to turn an unsupervised task into a supervised one. For example, ALOCC \cite{ref11} adds noisy data to positive samples fed to a generator network R, which is then trained to output the results by a discriminator network.

\begin{displaymath}\min _{\mathcal{R}} \max _{\mathcal{D}} \left(\mathbb{E}_{X \sim p_{t}}[\log (\mathcal{D}(X))]\right.
\left.+\mathbb{E}_{\tilde{X} \sim p_{t}+\mathcal{N}_{\sigma}}[\log (1-\mathcal{D}(\mathcal{R}(\tilde{X})))]\right)\end{displaymath}

Traditional machine vision foreign object detection methods, existing deep neural network-based foreign object detection algorithms, accuracy if required, then a large amount of labeled data, such as yolov5 \cite{ref7}, are required. However, if faced with industrial scenarios, the morphological distribution of negative samples is not stable compared to the large number of positive samples that appear, and it is difficult to collect in large quantities. Even if they are collected, the labor cost of labeling data and the time cost will rise sharply. So it is difficult to be practically applied. In contrast, most of the existing unsupervised algorithms based on deep neural networks use the middle layer of the network to calculate the distance. But the intermediate layer is highly abstracted and fuzzy, and cannot achieve high accuracy (pixel level). Alternatively, deep network modeling (e.g., similar schemes), where the network is very deep, causes a dramatic increase in training and inference time and data costs.
Methods of high relevance to this paper such as Noise2Noise [citation needed] are empirical risk minimization tasks that train the network to learn a zero-mean noise.
\begin{displaymath}z=\mathbb{E}_{y}\{y\}\end{displaymath}
\begin{displaymath}\underset{\theta}{\operatorname{argmin}} \mathbb{E}_{x}\left\{\mathbb{E}_{y \mid x}\left\{L\left(f_{\theta}(x), y\right)\right\}\right\}\end{displaymath}
For infinite data, the result is the same prediction formula when there is no noise, and for finite data, the variance is the average variance among the corruptions of the target. an exact P(noisy|clean), or P(clean), is not required, but can be obtained from the distribution of the data.

\section{Method}
\begin{figure}
\includegraphics[scale=0.115]{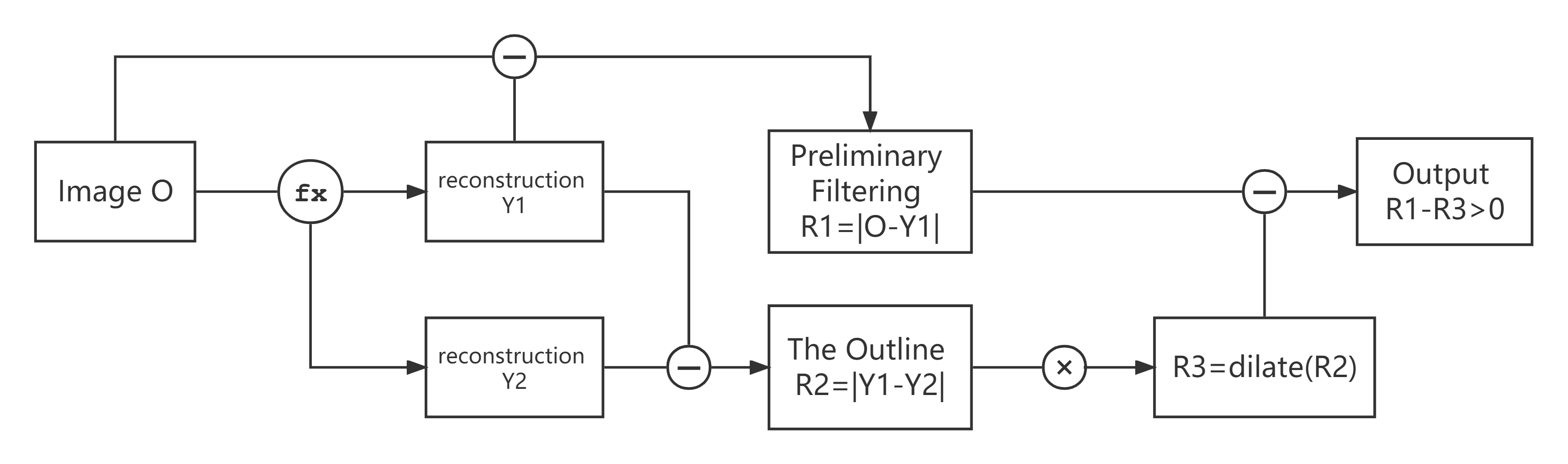}
\includegraphics[scale=0.12]{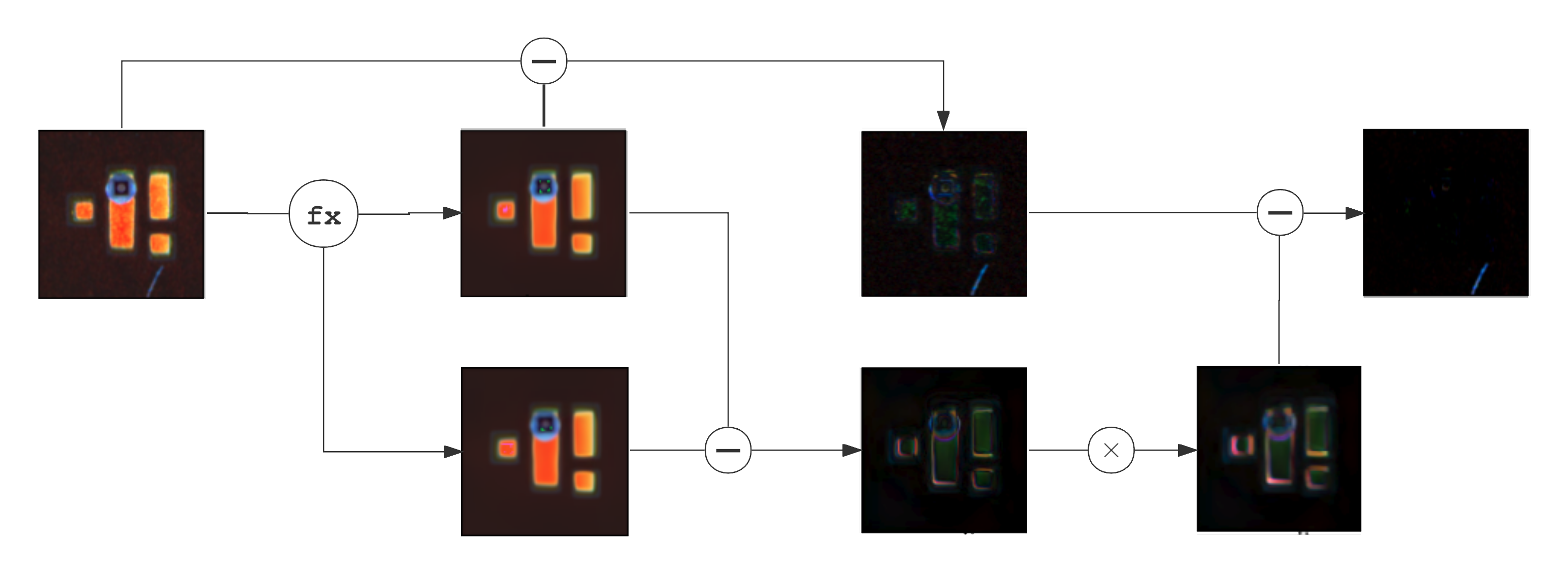}
\caption{A visualization of the method.The past method stops at R1, we go a step further and remove the edge errors from R1 to get a purer noise \& foreign matter image.} 
\label{fig:4}  
\end{figure}

In the training stage, we train the model using the method of training denoise self-encoders,which is usually treated as a regression model, using CNN:
\begin{displaymath}\underset{\theta}{\operatorname{argmin}} \mathbb{E}_{x}\left\{\mathbb{E}_{y \mid x}\left\{L\left(f_{\theta}(x), y\right)\right\}\right\}\end{displaymath}
where $fx$ is the CNN model, $x$ is the noisy sample, and $y$ is the noiseless sample. $l$ is the loss function.
The corresponding clean samples are then inferred from the noisy signal x.
\begin{displaymath}
\hat{x} \sim p\left(\hat{x} \mid y_{i}\right)
\end{displaymath}
We have the assumption that an image to be detected O = R + E + F + N is composed of
R is the reconstructed image, E is the edge error of R, F is foreign object, defects, and N is noise.
The design of the network can be quite arbitrary. Different perceptual fields and network depths can be decided depending on the size of the object to be detected. It is also possible for the network to search for a suitable feature scale (unet++)\cite{ref6} on its own. However, the following considerations should be noted: 1. Because it is a reconstruction task, there should not be a residual structure similar to resnet\cite{ref9}, to avoid the model being lazy and making a constant mapping directly. 2. One input, two outputs. The same input goes into two networks, which have a similar structure, the same training method, the same training data, but not the same parameters.

In the training process, we use L1Loss for the loss function. the optimizer uses the RMSprop method, the learning rate is set to 1e-4, and the momentum is set to 0.9.

Any image $O$ to be detected is fed into the training model and two reconstructed images $Y1$ and $Y2$ are produced.

Subtracting $O$ from the $Y1$ pixel level, taking the absolute value, produces $R1$ (foreign matter + contour gradient).

\begin{displaymath}R1=|O-Y1|\end{displaymath}

Subtracting the Y1 and Y2 pixel levels, taking the absolute value, produces $R2$ (the contour gradient).
\begin{displaymath}R2=|Y2-Y1|\end{displaymath}

A 3*3 dilate algorithm $D(X)$ is used for $R2$ to produce $R3$.

\centerline{$D(X) = $\{$a $|$ Ba$↑$X$\}}
\begin{displaymath}R3 = D(R2)\end{displaymath}

Subtract $R1$ from $R3$ and take 0 for the part less than 0 to produce the result $OUT$ (foreign object).
\begin{displaymath}OUT=|R3-R1|\end{displaymath}

We next count the brightness of the positive sample $OUT$ to get a histogram. 
\begin{displaymath}\sum_{i=1}^{n}{} \sum_{j=1}^{n}{OUT_{ij}}\end{displaymath}

We can set the desired confidence threshold based on this histogram.

\section{Experimental, comparative data}

So far we have experimented and found that a 4+4 layer network and a 3+3 layer network obtained better results when choosing images at the 200*200 pixel scale. We have tried a 2*4+4 layer twin network and a 4+4 and 4+4 dual flow network. Neither was found to work as well as the above structures.

We also experimented with comparing the more recent and popular MAE\cite{ref5} method in the course of training data. As the graph shows, the best results are obtained when the percentage of patches is 0, i.e. without patches. As the epoch increases, the overall accuracy moves towards 0patch.
\begin{figure}[htb]
\subfigure
[Different scales of Patches on the accuracy of model reduction.]
{\includegraphics[scale=0.4]{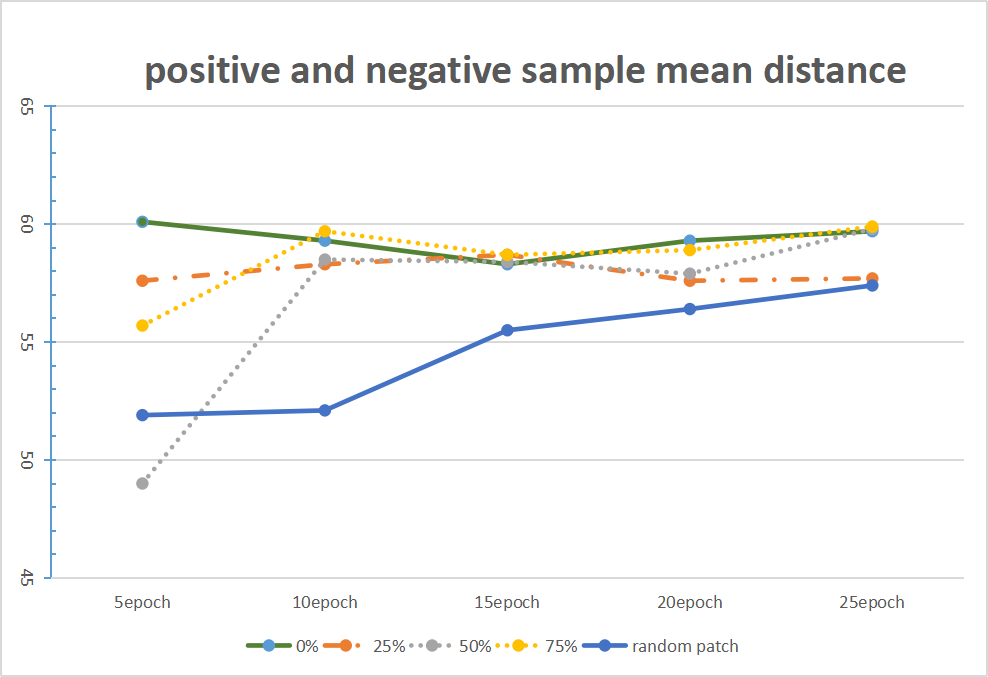}}
\subfigure
[The performance of the algorithm on our own test set, where the green part is the positive sample distribution and the red part is the negative sample distribution.]
{\includegraphics[scale=0.15]{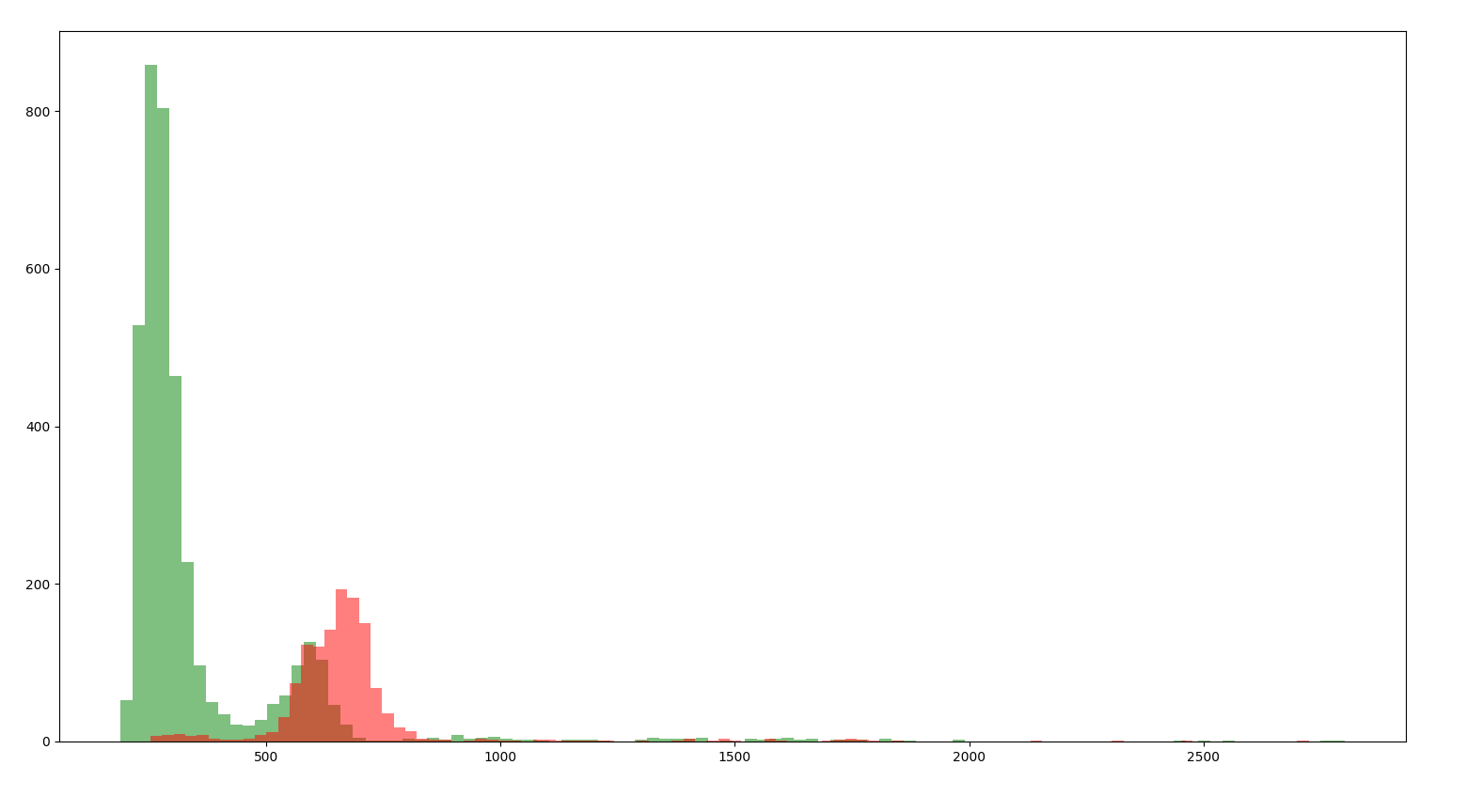}}
\caption{shows some training test data of the model on our private dataset.}
\label{fig:5}  
\end{figure}

\begin{figure}[htb]
\subfigure
{\includegraphics[scale=0.2]{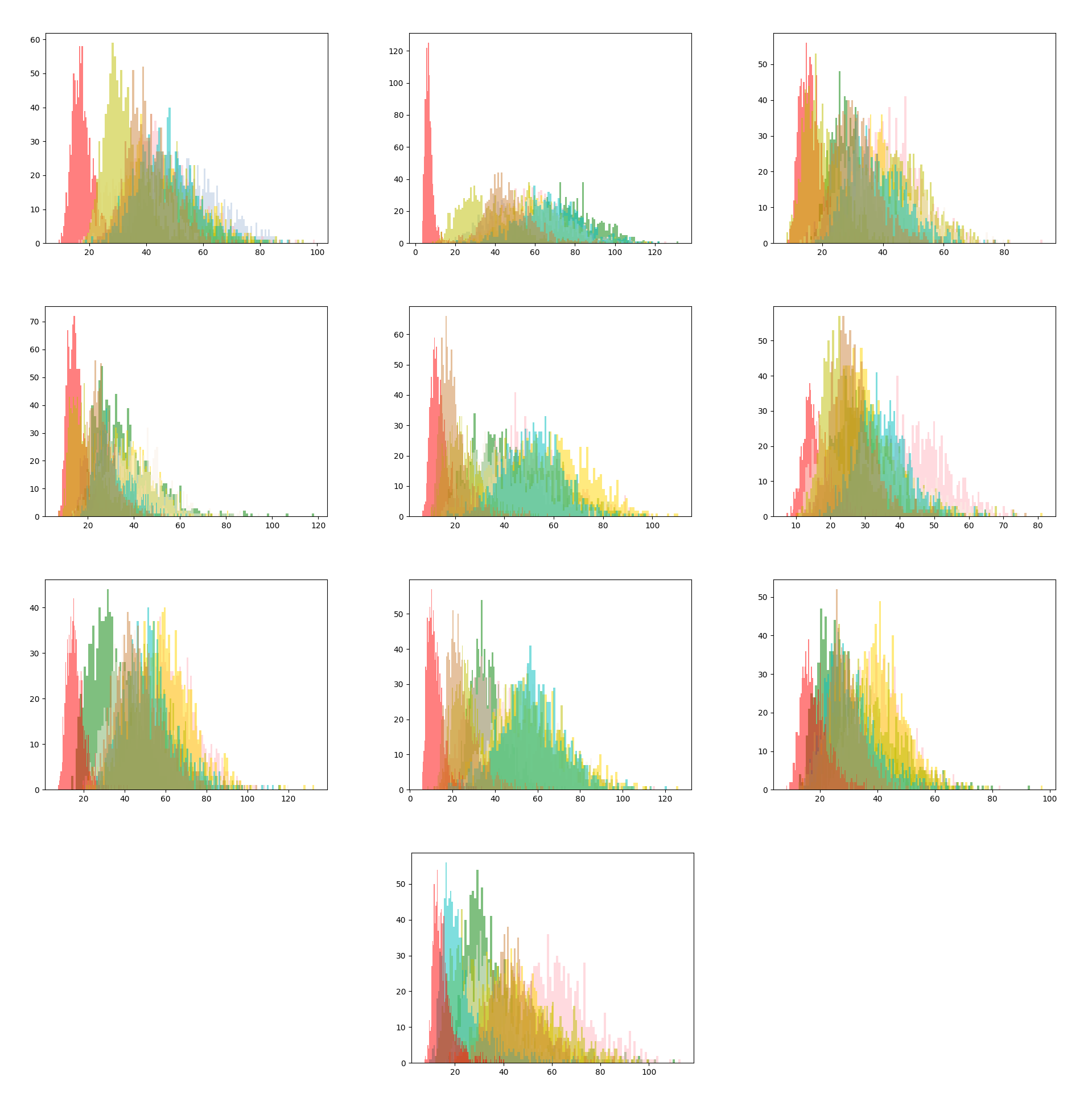}}
\caption{shows the experimental results of our algorithm on the Mnist dataset. The ten graphs represent the effect of the model trained for ten numbers from 0-9. Take the first graph as an example, it represents the training process, we only use data 0 in the training set to train the model, and then go to the test set, input all test data from 0-9 into the model, output the result, and get the statistical histogram. The red color represents the distribution of the training data of the model, and the other colors represent the distribution of the remaining numbers. From this graph, it is obvious that our algorithm has advanced performance.}
\label{fig:6}  
\end{figure}

\begin{figure}[htb]
\subfigure
{\includegraphics[scale=0.6]{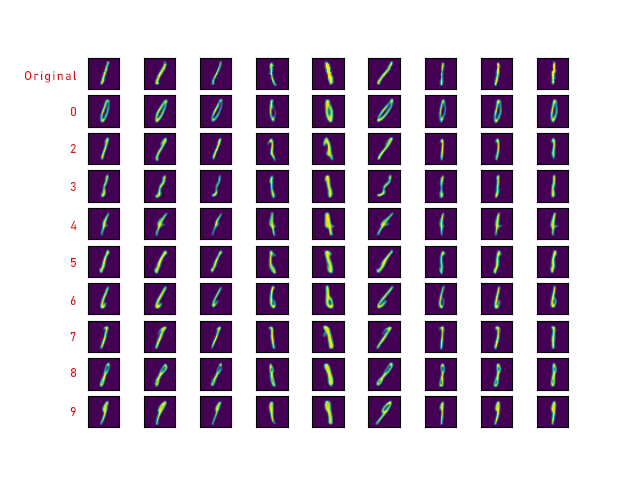}}
\caption{shows the experimental results of our algorithm on the Mnist dataset. The ten graphs represent the effect of the model trained for ten numbers from 0-9. Take the first graph as an example, it represents the training process, we only use data 0 in the training set to train the model, and then go to the test set, input all test data from 0-9 into the model, output the result, and get the statistical histogram. The red color represents the distribution of the training data of the model, and the other colors represent the distribution of the remaining numbers. From this graph, it is obvious that our algorithm has advanced performance.}
\label{fig:7}  
\end{figure}
\section{Follow up}

So far, it has been possible to get a relatively pure image containing only noise + foreign matter. However, as you can see, as long as the noise is still present, the judgement is not sufficient. Further work can be done to analyse the noise and to separate it from the foreign matter. The accuracy can then be further improved.

\begin{ack}
Use unnumbered first level headings for the acknowledgments. All acknowledgments
go at the end of the paper before the list of references. Moreover, you are required to declare
funding (financial activities supporting the submitted work) and competing interests (related financial activities outside the submitted work).
More information about this disclosure can be found at: \url{https://neurips.cc/Conferences/2021/PaperInformation/FundingDisclosure}.

Do {\bf not} include this section in the anonymized submission, only in the final paper. You can use the \texttt{ack} environment provided in the style file to autmoatically hide this section in the anonymized submission.
\end{ack}

\medskip

{
\small

}


\end{document}